\documentclass{article} 
\usepackage{iclr2025_conference,times}

\usepackage{url}
\usepackage{amsmath, amssymb, tikz, algorithm, algorithmic}
\usetikzlibrary{matrix,shapes,arrows,positioning}
\usepackage{geometry}
\usepackage{booktabs, multirow}
\usepackage{graphicx}
\usepackage{caption}
\usepackage{subcaption}
\usepackage{hyperref}
\usepackage{arydshln}
\usepackage{makecell}
\usepackage{colortbl}

\makeatletter
\renewcommand\section{\@startsection{section}{1}{\z@}%
  {-1.5ex \@plus -1ex \@minus -.2ex}%
  {1.5ex \@plus .3ex}%
  {\normalfont\large\bfseries}}
\renewcommand\subsection{\@startsection{subsection}{2}{\z@}%
  {-1.25ex\@plus -1ex \@minus -.2ex}%
  {1.25ex \@plus .25ex}%
  {\normalfont\normalsize\bfseries}}
\makeatother

\renewenvironment{abstract}{
    \vskip.075in\centerline{\large\bfseries Abstract}\vspace{0.5ex} 
    \begin{quote}
    }
    {\end{quote}\vskip 1ex}

\usepackage{authblk}  

\renewcommand\footnotemark{}  

\author{
Rasoul Shafipour\textsuperscript{1},
David Harrison\textsuperscript{1},
Maxwell Horton\textsuperscript{1}, 
Jeffrey Marker\textsuperscript{1},
Houman Bedayat\textsuperscript{1}, \\
Sachin Mehta\textsuperscript{2},
Mohammad Rastegari\textsuperscript{2}, 
Mahyar Najibi\textsuperscript{1}, 
Saman Naderiparizi\textsuperscript{1} \\
\vspace{10pt}
\textsuperscript{1}Apple \hspace{10pt}
\textsuperscript{2}Meta 
}
\thanks{Corresponding authors: \texttt{\{rshafipour, najibi, snaderiparizi\}@apple.com}. Work done by S. Mehta and M. Rastegari while at Apple.}

\newcommand{\draftonly}[1]{#1} 
\renewcommand{\draftonly}[1]{}

\iclrfinalcopy 

\begin{document}

\renewcommand{\headrulewidth}{0pt}  
\title{\normalfont{SeedLM: Compressing LLM Weights into Seeds of \\ Pseudo-Random Generators}}

\begin{center}
\maketitle
\end{center}

\begin{abstract}
Large Language Models (LLMs) have transformed natural language processing, but face significant challenges in widespread deployment due to their high runtime cost. In this paper, we introduce SeedLM, a novel post-training compression method that uses seeds of pseudo-random generators to encode and compress model weights. Specifically, for each block of weights, we find a seed that is fed into a Linear Feedback Shift Register (LFSR) during inference to efficiently generate a random matrix. This matrix is then linearly combined with compressed coefficients to reconstruct the weight block. SeedLM reduces memory access and leverages idle compute cycles during inference, effectively speeding up memory-bound tasks by trading compute for fewer memory accesses. Unlike state-of-the-art compression methods that rely on calibration data, our approach is data-free and generalizes well across diverse tasks.  Our experiments with Llama 3 70B, which is particularly challenging to compress, show that SeedLM achieves significantly better zero-shot accuracy retention at 4- and 3-bit  than state-of-the-art techniques, while maintaining performance comparable to FP16 baselines. Additionally, FPGA-based tests demonstrate that 4-bit SeedLM, as model size increases to 70B, approaches a 4x speed-up over an FP16 Llama 2/3 baseline. 

\end{abstract}

\section{Introduction}\label{S:Introcution}

Large Language Models (LLMs) have demonstrated impressive performance across numerous benchmarks \citep{achiam2023gpt,touvron2023llama}. However, the practical deployment of these models often encounters limitations due to substantial memory transfer requirements. This issue is especially pronounced during autoregressive generation, which is primarily memory-bound and takes the majority of the inference time~\citep{lee2024cost}. In contrast, operations like 8-bit integer multiplication performed at 45nm 0.9V are demonstrated to be over 800x more energy-efficient than reading the same 8 bits from DRAM~\citep{horowitz2014computing}. In this paper, we explore the following question: Can we trade a reasonable increase in compute for a reduction in memory accesses? A positive answer here not only transforms energy-intensive memory access operations into more energy-efficient compute operations but also alleviates the memory bandwidth limitations that pose a significant bottleneck during LLM inference.

Post-training weight compression is an effective method to reduce the size of pretrained LLMs, making them suitable for on-device execution or reducing power consumption through fewer memory reads. Current state-of-the-art techniques for compressing weights typically require calibration data and involve meticulously adjusting the weights to ensure that the learned knowledge is retained.

We introduce SeedLM, a simple yet effective compression technique which can compress weights to 3-4 bits with minimal accuracy loss. SeedLM is an innovative method for compressing the weights of LLMs 
by projecting weight blocks into pseudo-random projection basis sets. By finding the optimal seeds to generate these pseudo-random projections per weight block, SeedLM ensures a low compression error and consequently, maintains the accuracy of the original model. Our approach only requires storing the seed and few projection coefficients instead of all the weight values to reconstruct high dimensional weight blocks.

As a result, SeedLM significantly reduces the memory footprint required for operating large-scale models during inference. To generate pseudo-random matrix blocks given a seed, we leverage Linear Feedback Shift Register (LFSR) hardware blocks that are widely used in applications such as cryptography, communication, and error detection \citep{gaitonde1988tutorial,zeng2013reconfigurable,xiang2016low}. LFSRs can be efficiently implemented in the silicon with minimal energy and area footprint.

\begin{figure}[t]
\centering
 \includegraphics[width=\linewidth]{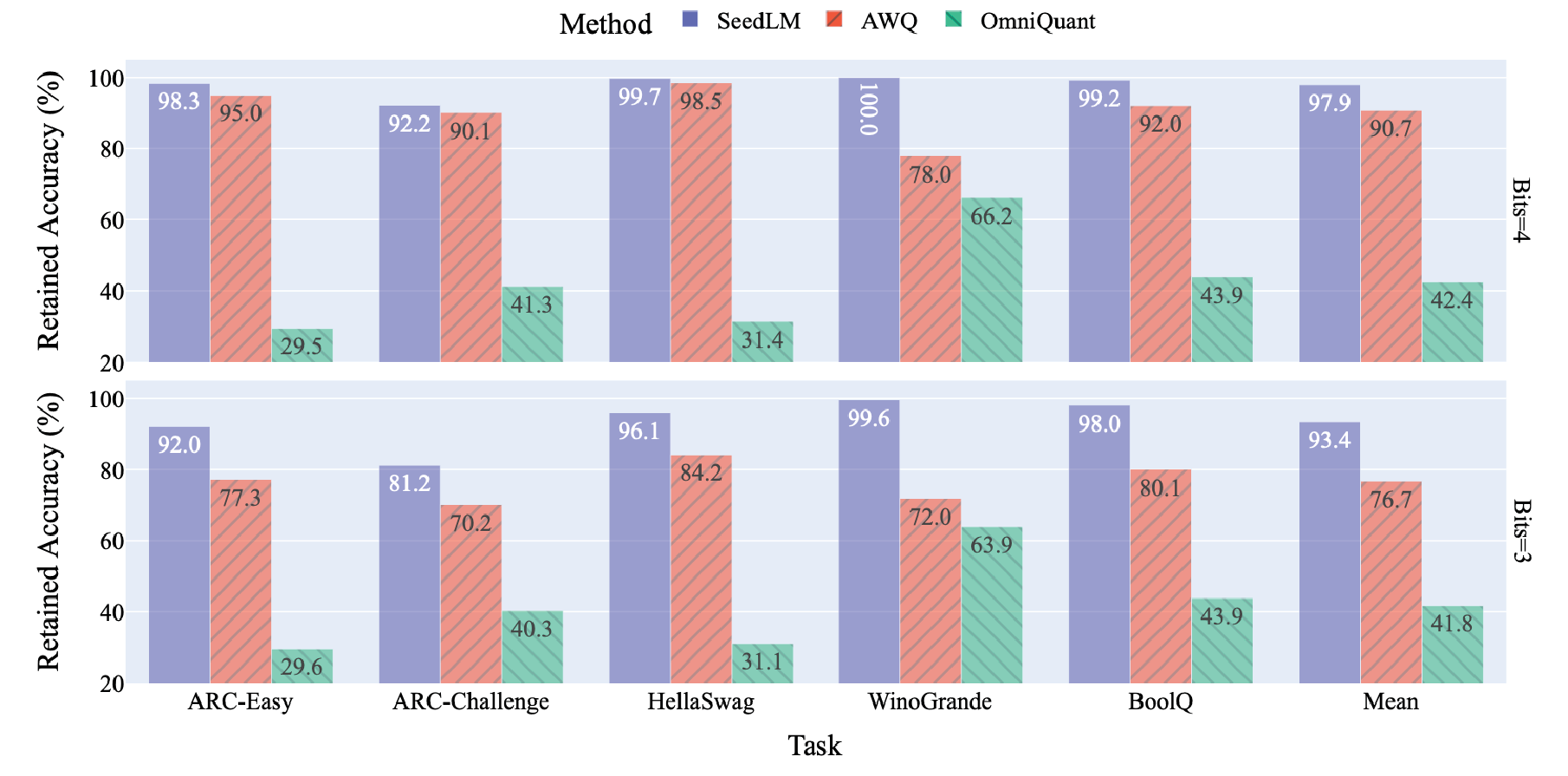}
\caption{Retained zero-shot accuracy across a variety of tasks and compression methods, compared to the FP16 Llama 3 70B model. The top row shows data for 4-bit compression, while the bottom row shows data for 3-bit compression. We compare the performance of SeedLM, AWQ, and OmniQuant across the ARC-Easy, ARC-Challenge, HellaSwag, WinoGrande, and BoolQ tasks. While being completely data-free, SeedLM outperforms state-of-the-art weight quantization methods that rely on a calibration dataset.}
\label{fig:accuracy_compare}
\end{figure}

Figure~\ref{fig:accuracy_compare} shows the Retained Accuracy (\%), which is the ratio of the compressed model's accuracy to the full-precision FP16 model's accuracy, on the Llama 3 70B model. As illustrated, SeedLM retains approximately 97.9\% of the zero-shot accuracy on average across various tasks in a data-free setting, using 4 bits per weight element (see Section~\ref{ss:exp_acc} for more details). It also consistently outperforms state-of-the-art 3-bit and 4-bit compression techniques that rely on calibration data.  To the best of our knowledge, this is the first time nearly identical accuracy has been achieved with 4-bit compression on LLMs without data, using a deterministic offline algorithm.

In summary, our key contributions are:
\begin{itemize}
  \item We demonstrate, for the first time, how LFSR hardware blocks can be leveraged to trade increased compute for reduced memory accesses.
  \item We show the first instance of achieving nearly identical accuracy with 4-bit quantization without data, using a deterministic offline algorithm.
  \item We demonstrate an effective solution to find the optimized seed for the LFSR modules while maximizing compression ratio.
  \item We prototype SeedLM on an FPGA (Field-Programmable Gate Array) and demonstrate its efficacy in reducing the inference latency with custom hardware.
\end{itemize}

\section{Related Work}  \label{S:Related}
Significant research has been conducted in model compression for LLMs, a critical approach for reducing both the memory footprint and computational demands of these models.  In this section, we highlight some of the most relevant techniques from prior work.

\textbf{Compression With Random Basis:} Recent works have demonstrated that neural networks can be decomposed into random number generator seeds and weight coefficients. In PRANC \citep{Nooralinejad2022PRANCPR}, full networks are compressed by orders of magnitude to improve storage and transmission efficiency. LoRA \citep{Hu2021LoRALA} compresses the weights by injecting trainable rank decomposition matrices into each layer of the network. NOLA \citep{Koohpayegani2023NOLACL} builds upon LoRA by compressing the low-rank matrices through a linear combination of random basis vectors, further reducing memory and computational overhead. 


Our work (SeedLM) is conceptually similar in that we compress networks using a random basis. However, a key distinction from NOLA is that they do not utilize lightweight pseudo-random generator modules. Thus, they haven't demonstrated the ability to efficiently generate weights on-the-fly by colocating weight generation with computation. Additionally, unlike SeedLM, they don’t find an optimal seed but instead rely on a random seed. These models also use much larger basis ranks applied globally rather than per-block, leading to significantly more operations per parameter to preserve accuracy, making them computationally infeasible for LLM inference.

\textbf{Data-Free Post-Training Compression:} A few previous works have explored data-free post-training compression \citep{Nagel2019DataFreeQT,Horton2020LayerWiseDC,Nunez2023LCSLC}. Such works are capable of producing a compressed model after training without the need for calibration data. They usually apply quantization or pruning techniques to obtain a smaller model. Similarly, SeedLM does not require any data for model compression. This is in contrast to most recent works on LLM compression, which require calibration data.

There are more computationally expensive methods for data-free compression that involve generating data from a teacher model and performing distillation \citep{Lopes2017DataFreeKD,Gou2020KnowledgeDA}. These techniques are applied to LLMs as demonstrated in \citet{Liu2023LLMQATDQ}.

\textbf{Post-Training Compression with Calibration Data:} An early example of post-training quantization with calibration data is found in \citet{Nagel2020UpOD}, where activation statistics are used to decide whether to round quantized values up or down. The cost of post-training quantization is a small fraction of the total training cost.

Recently, calibration data has been leveraged for post-training compression in LLMs. AWQ \citep{lin2024awq} rescales salient weights before compression using activation statistics. In QuIP\# \citep{tseng2024quip} and GPTQ \citep{frantar2022gptq}, Hessian analysis of calibration data helps make rounding decisions during quantization. SpQR \citep{dettmers2023spqr} retains outliers during quantization to preserve accuracy. OmniQuant \citep{shao2023omniquant} employs weight clipping and other transformations to maintain accuracy. Additive Quantization \citep{egiazarian2024extreme} learns a codebook for performing additive quantization. In our study, we used AWQ, QuIP\#, and OmniQuant as our main baselines because they avoid costly training while delivering strong results.

\textbf{Training-Aware Compression:} Compressing the model during the training process has the disadvantage of fixing the compression method and parameters beforehand, but usually offers better accuracy. An overview of quantization-aware training is provided in \citet{Jacob2017QuantizationAT,nagel2022overcoming, xi2023training}, while pruning techniques are discussed in \citet{AlizadehVahid2023LLMIA,Sun2023ASA,Kusupati2020SoftTW,LeCun1989OptimalBD}. In \citet{rouhani2023microscaling}, stable training and post-training quantization are demonstrated using hardware-friendly 4-8 bit weights, activations, and gradients with minimal accuracy loss by utilizing micro-scaled data formats.

In this work, we focus on post-training weight compression with SeedLM, and in the following sections, we outline our methodology and present the results.

\section{Methodology}  \label{S:Methods}
In this section, we introduce SeedLM, our method for compressing the weights of LLMs by using seeds from a pseudo-random generator. Initially, each weight matrix is segmented into blocks of $C$ contiguous elements. Representing each block as a vector $\mathbf{w} \in \mathbb{R}^C$, we approximate it as a linear combination of columns from a matrix $\mathbf{U} \in \mathbb{R}^{C \times P}$. This matrix $\mathbf{U}$ is constructed using a pseudo-random generator given a seed specifically selected to generate a subspace that most effectively reconstructs $\mathbf{w}$ linearly.

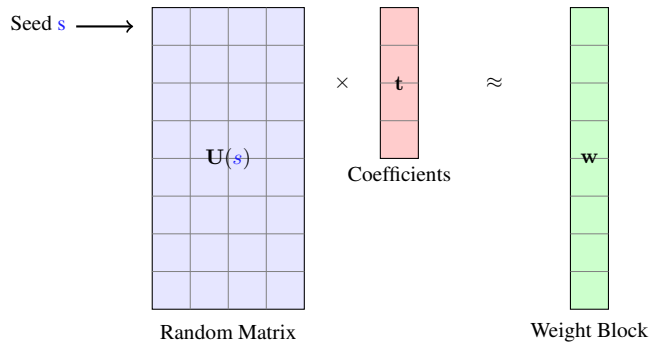
\begin{figure}[htbp]
\centering
\begin{tikzpicture}[scale=1.0, every node/.style={scale=0.8}]
    \draw[fill=blue!10] (0,0) rectangle (2,4);  
    \foreach \x in {0.5,1,1.5} {
        \draw[gray, very thin] (\x,0) -- (\x,4);
    }
    \foreach \y in {0.5,1,...,3.5} {
        \draw[gray, very thin] (0,\y) -- (2,\y);
    }
    \node at (1,2) {$\mathbf{U}(\textcolor{blue!90}{s})$};
    
    \draw[fill=red!20] (3,2) rectangle (3.5,4);
    \foreach \y in {2.5,3,3.5} {
        \draw[gray, very thin] (3,\y) -- (3.5,\y);
    }
    \node at (3.25,3) {$\mathbf{t}$};
    
    \node at (2.5,3) {$\times$};
    
    \node at (4.5,3) {$\approx$};
    
    \draw[fill=green!20] (5.5,0) rectangle (6,4);
    \foreach \y in {0.5,1,...,3.5} {
        \draw[gray, very thin] (5.5,\y) -- (6,\y);
    }
    \node at (5.75,2) {$\mathbf{w}$};
    
    \node at (1,-0.3) {Random Matrix};
    \node at (3.25,1.8) {Coefficients};
    \node at (5.75,-0.3) {Weight Block};
    
    \draw [->,thick] (-1,3.75) -- (-0.25,3.75);
    \node[align=center] at (-1.5,3.80) {Seed \textcolor{blue!90}{s}};
\end{tikzpicture}
\caption{Compression of weights using pseudo-random generated matrices.}
\label{fig:weight-compression}
\end{figure}

Figure~\ref{fig:weight-compression} illustrates this setup. Our primary goal is to find the optimal seed, $s$, and coefficient vector, $\mathbf{t} \in \mathbb{R}^{P}$, that minimize the reconstruction error between the original and the approximated weights. For this reconstruction, only the seed and the coefficients are stored. In the following subsection, we first outline a mechanism to efficiently generate 
$\mathbf{U}$ using a $K$-bit seed in our Linear Feedback Shift Register (LFSR) framework. We will then discuss the methodologies employed to determine 
$s$ and $\mathbf{t}$.

\subsection{Linear Feedback Shift Register (LFSR)}
\label{SS:LFSR}
A Linear Feedback Shift Register (LFSR) is a simple yet effective type of shift register, ideal for generating pseudo-random binary sequences. The primary advantages of LFSRs in hardware include cost-effectiveness and minimal resource consumption due to their straightforward implementation with basic flip-flops and XOR gates. This simplicity facilitates rapid and efficient sequence generation, which is integral to our compression technique.

An LFSR operation can be characterized by its length $K$ (which determines the number of bits in its shift register) and its feedback polynomial. To generate next pseudo-random number in the sequence, each bit in the register is first shifted to the next position. Then, the new bit entering the register is calculated as a linear combination of certain bits of the current state as specified by the feedback polynomial, typically implemented by XOR operations. Mathematically, the new bit $x_{n+1}$ generated by the LFSR can be expressed as:
\[
x_{n+1} = \sum_{i=0}^{K-1} \alpha_{i} \cdot x_{n+i-K+1} \mod 2,
\]
where $K\!\geq\!2$ and $\alpha_0, \ldots, \alpha_{K}$ are the binary coefficients that define the feedback polynomial, with each $\alpha_j$ determining whether the bit $x_j$ is selected or not.

The state transition in the LFSR can be described as follows: if the current state is represented by the bits $x_n, x_{n-1}, \ldots, x_{n-K+1}$, then after the shift, the new state will be $x_{n+1}, x_n, \ldots, x_{n-K+2}$. This transition reflects the shift of every bit to the right by one position, with the new bit $x_{n+1}$ entering at the leftmost position.
Given its finite state nature, an LFSR will eventually enter a repeating cycle, suggesting an asymptotic uniform distribution over this cycle. An LFSR can cycle through at most
$2^K\!-\!1$ states, excluding the all-zero state.

A key goal when designing an LFSR is to guarantee a maximal-length sequence. This ensures that the LFSR will produce the longest possible sequence of non-repeating states before repeating. Intuitively, this means the LFSR will cycle through every possible state (except the all-zero state), maximizing the number of distinct pseudo-random values generated. To achieve this maximal-length property, the feedback polynomial must be primitive over the Galois field GF(2). In simple terms, a primitive polynomial ensures that the LFSR explores all $2^K\!-\!1$ states without prematurely entering a repeating cycle. 
 
For our experiments, this means a fixed set of coefficients $\{\alpha_j: 0\leq j \leq K\!-\!1\}$ that is hard-wired, ensuring maximal length if and only if it avoids the all-zero state, in which it stays in zero. Refer to Section~\ref{appendix:lfsr_coeff} for the indexed $j$ coefficients used for each $K$ where $\alpha_j$ equals one; all other coefficients are zero. For a comprehensive understanding of LFSRs and their properties, see \citep{bhattacharjee2022search}.

\begin{figure}[ht]
    \centering
    \begin{tikzpicture}
    \def\radius{1.5cm} 
    \node[circle,draw] (6) at (115.68:\radius) {6};
    \node[circle,draw] (7) at (167.1:\radius) {7};
    \node[circle,draw] (3) at (218.52:\radius) {3};
    \node[circle,draw] (1) at (270:\radius) {1};
    \node[circle,draw, very thick] (4) at (321.42:\radius) {4};
    \node[circle,draw] (2) at (12.84:\radius) {2};
    \node[circle,draw] (5) at (64.26:\radius) {5}; 

    \draw[->] (4) -- (2);
    \draw[->] (2) -- (5);
    \draw[->] (5) -- (6);
    \draw[->] (6) -- (7);
    \draw[->] (7) -- (3);
    \draw[->] (3) -- (1);
    \draw[->] (1) -- (4);

    \node[right=5cm of 7] (matrixlabel) {\(\mathbf{V}(4) = \)};
    \matrix[matrix of math nodes, right=0.1cm of matrixlabel, left delimiter={[},right delimiter={]}] (m)
    {
      2 & 5  \\
      6 & 7 \\
      3 & 1 \\
      4 & 2 \\
    };

\end{tikzpicture}
    \caption{Illustration of the state sequence for a $K\!=\!3$ LFSR with all possible states with the feedback polynomial defined in Table~\ref{appendix:lfsr_coeff}. The matrix \(\mathbf{V}(4)\) starts filling with the value generated one cycle after the seed state $s\!=\!4$, which is highlighted with a thick circle.   
    }
    \label{fig:lfsr_example}
\end{figure}
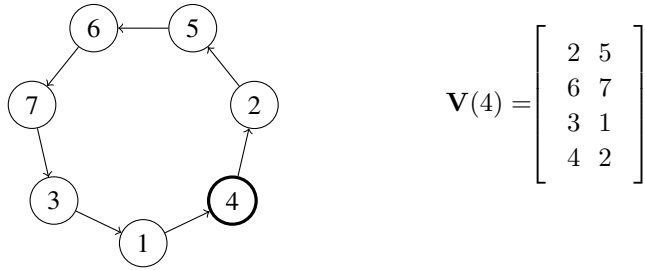

To optimize the efficiency of generating random matrices through an LFSR, for a fixed length $K$ and a set of coefficients $\{\alpha_j\}$, we cache all the 
$2^K\!-\!1$ states that sequentially follow each other -- each state uniquely determined by its preceding state along with 
$K$ and $\{\alpha_j\}$. This cache allows us to extract an arbitrarily sized random matrix from the sequence given a random seed $s$, where the matrix begins to fill up starting from the first value generated by the LFSR after the seed, not the seed itself. We can cycle through these states to generate matrices of any desired size. For an illustration, refer to Figure~\ref{fig:lfsr_example}. With $K\!=\!16$ and a maximal length LFSR, all states will occupy approximately $(2^{16}\!-\!1) \times 2 \text{ Bytes} \approx 130$KB of memory, which is negligible. This setup ensures a highly efficient and scalable mechanism for generating the necessary random matrices for our compression technique. $K$ is a hyper-parameter of our method which we will elaborate on in Section~\ref{ss:design-space}.

\subsection{Weight Compression Using Pseudo-Random Generators} \label{SS:Weight_comp}
Building on the foundations laid by the LFSR mechanisms, our methodology seeks to represent a block of data $\mathbf{w} \in \mathbb{R}^{C}$ using the decomposition $\mathbf{U}(s) \mathbf{t}$. Here, $\mathbf{U}(s) \in \mathbb{R}^{C \times P}$ is a random matrix derived from a $K$-bit maximal-length LFSR, ensuring the full sequence of possible states is used. As the output of the LFSR is in the range of $[1, 2^K\!-\!1]$, which are unsigned integers, we normalize them to ensure they fall within the range of $[-1,1]$, enhancing their expressiveness. Importantly, the seed $s$ used to initialize the LFSR is non-zero to avoid the degenerate all-zero state. More specifically, to construct $\mathbf{U}(s)$, we first generate a matrix $\mathbf{V}(s)$ using an LFSR seeded by $s$ as illustrated in Figure~\ref{fig:lfsr_example}. This matrix initially contains integers and undergoes the following normalization to ensure that its components lie between $[-1,1]$:

\begin{equation} \label{eq:LFSR_U}
  \mathbf{U}(s) = \frac{1}{2^{K-1} - 1} \left(\mathbf{V}(s) - 2^{K-1} \mathbf{1}\right),  
\end{equation}

where $\mathbf{1} \in \mathbb{R}^{C \times P}$ represents a matrix of all ones, and $K$ is the LFSR bit length. To determine the optimal seed and coefficients, we solve the following optimization problem:
\begin{equation} \label{eq:opt} 
\begin{aligned}
& \underset{s, \mathbf{t}}{\text{minimize}} \quad \lVert \mathbf{w} - \mathbf{U}(s) \mathbf{t} \rVert_2^2, \quad \text{subject to:} \quad s \in \{1, \dots, 2^K - 1\}, \quad \text{and}  \quad \mathbf{t} \in \mathcal{Q},
\end{aligned}
\end{equation}
where $\lVert \cdot \rVert$ denotes the Euclidean norm, and $\mathcal{Q}$ represents the set of valid quantized values. The objective is to identify an optimal seed $s^*$ and coefficients $\mathbf{t}^*$ that most effectively approximate $\mathbf{w}$. This problem is NP-hard due to the discrete nature of the seed selection and the quantization of coefficients since it involves searching through a combinatorially large set of possibilities. Next, we describe the design choices and heuristics used to solve the problem.

The quantization scheme for the vector $\mathbf{t}$ plays a critical role in balancing reconstruction accuracy with bit efficiency, adhering to our bit budget constraints. We represent each element of $\mathbf{t}$ as a 4-bit two's complement integer, paired with a shared 4-bit exponent. This configuration allows us to capture a broad dynamic range of values within the interval $[-8 \times 2^{-8}, 7 \times 2^{7}]$. The shared 4-bit exponent further extends the dynamic range, enabling representation across several orders of magnitude. We have specifically chosen for the exponent to be a power of two, which is hardware-friendly and can be efficiently implemented using simple shift operations in digital circuits; see \citep{darvish2020pushing}. By combining the 4-bit integer with the shared exponent, each quantized element is expressed as $t_i = q_i \times 2^e$, where $q_i$ is the 4-bit two's complement integer and $e$ is the shared exponent. The shared exponent $e$ is selected as:
\[
e = \max_i \left\lfloor \log_2 \left( \left| t_i \right| \right) \right\rfloor,
\]

where $\left| t_i \right|$ is the absolute value of each element of $\mathbf{t}$. After determining $e$, each $t_i$ is scaled by dividing it by $2^e$, and then quantized to a 4-bit two's complement integer to derive $q_i$.

\subsection{Approximation Approach}

Looking back at the optimization problem in Eq.~\ref{eq:opt}, while the unconstrained case admits a closed-form solution given by $\mathbf{U}(s)^{\dagger} \mathbf{w}$, where $\mathbf{U}(s)^{\dagger}$ denotes the Moore-Penrose pseudo-inverse of $\mathbf{U}(s)$, our discrete constraints convert it to an NP-hard optimization problem. Hence, to solve Eq.~\ref{eq:opt}, we employ an approximate heuristic approach that involves the following steps:

\begin{enumerate}
    \item Generate $N = 2^K\!-\!1$ random matrices $\{ \mathbf{U}(s_1), \mathbf{U}(s_2), \dots, \mathbf{U}(s_{N}) \}$, each of size $C \times P$, with an LFSR of length $K$ based on Eq.~\ref{eq:LFSR_U} and $s_j\!:=\!j$.
    
    \item For each matrix $\mathbf{U}(s_j)$, project the vector $\mathbf{w}$ onto the subspace spanned by $\mathbf{U}(s_j)$:
    \begin{equation} \label{eq:pseudo} \notag
    \mathbf{t}_j = \mathbf{U}(s_j)^{\dagger} \mathbf{w}.
    \end{equation}
    
    \item Quantize $\mathbf{t}_j$ to obtain the vector $\hat{\mathbf{t}}_j$ using 4-bit integers and a 4-bit shared exponent $e_j$.
    
    \item Compute the reconstruction error for each pair $(\mathbf{U}(s_j), \hat{\mathbf{t}}_j)$ as follows:
    \[
    \epsilon_j = \| \mathbf{w} - \mathbf{U}(s_j) \hat{\mathbf{t}}_j \|_2^2.
    \]
    
    \item Select the pair $(\hat{s^*}, \hat{\mathbf{t}^*})$ that minimizes the reconstruction error:
    \begin{equation} \label{eq:min_error}
    (\hat{s^*}, \hat{\mathbf{t}^*}) = \arg\min_{s_j, \hat{\mathbf{t}}_j} \epsilon_j.
    \end{equation}
\end{enumerate}
Our heuristic algorithm leverages randomness to explore multiple subspaces and selects the one that best approximates $\mathbf{w}$ under the given constraints.

\begin{minipage}{0.33\textwidth}
In summary, we apply the above heuristics across all weight blocks in parallel to find the seeds and coefficients that minimize the reconstruction error. To enhance computational efficiency, we precompute and cache the pseudo-inverse matrices for all seeds and perform steps 2--5 in parallel across all blocks.  This process is summarized in Algorithm~\ref{Alg}. The inner loop can also be parallelized, and for the chosen $C$, $P$, and $K$ in the following design space exploration, the pseudo-inverses take up around 6.3MB at most, which is negligible.
\end{minipage}
\hfill
\begin{minipage}{0.64\textwidth}
\begin{algorithm}[H]
\caption{Seed and Coefficient Selection for a Weight Block}
\begin{algorithmic}[1] \label{Alg}
\REQUIRE $\mathbf{w} \in \mathbb{R}^C$, $\{\mathbf{U}(j)^{\dagger}\}_{j=1}^{2^K\!-\!1}$
\STATE $\hat{s^*} \leftarrow \text{null}, \hat{\mathbf{t}}^* \leftarrow \text{null}$
\STATE $\text{min\_norm} \leftarrow \infty$
\FOR{$j = 1$ \TO $2^K - 1$}
    \STATE $\mathbf{t} \leftarrow q(\mathbf{U}(j)^{\dagger} \mathbf{w})$, where $q(\cdot)$  quantizes its arguments to the set $\mathcal{Q}$
    \STATE $\text{norm} \leftarrow \lVert \mathbf{w} - \mathbf{U}(j) \cdot \mathbf{t} \rVert$
    \IF{$\text{norm} < \text{min\_norm}$}
        \STATE $\text{min\_norm} \leftarrow \text{norm}$
        \STATE $\hat{s^*} \leftarrow s$
        \STATE $\hat{\mathbf{t}^*} \leftarrow \mathbf{t}$
    \ENDIF
\ENDFOR
\RETURN $\hat{s^*}$, and $\hat{\mathbf{t}^*}$
\end{algorithmic}
\end{algorithm}
\end{minipage}

\subsection{Design Space Exploration} \label{ss:design-space}
The minimum reconstruction error obtained from Eq.~\ref{eq:min_error} depends on the block size \(C\), the latent dimension \(P\), and the LFSR length \(K\). Here, we explore how we select the optimal configuration for an \(M\)-bit compression. First, let's examine the total number of bits required to store a SeedLM block of \(C\) elements, which consists of the following:
 \begin{itemize}
 \item $K$ bits to index the selected random seed $\hat{s^*}$ to generate matrix $\mathbf{U}(\hat{s^*})$ among the $N\!=2^{k}\!-\!1$ candidates.
 \item 4 bits to store the shared exponent $e$.
 \item $4P$ bits to store the quantized vector $\hat{\mathbf{t}^*}$ ($P$ elements each requiring 4 bits). \end{itemize}

So, the effective bit per element is a function of hyper-parameters $K$, $C$, and $P$. In particular, for an $M$-bit compression, we have the bit budget per element as 
\begin{equation}
    \label{eq:bit_per_element}
    M = \frac{K + 4 + 4P}{C}.
\end{equation}

To determine the optimal configuration for a given bit budget per element, $M$, we evaluate the reconstruction accuracy of our method in the search space. Specifically, we explore how a standard normal Gaussian vector $\mathbf{w}$ can be approximated using any combination of valid hyperparameters given the bit budget $M$. While assuming a Gaussian distribution may have its limitations, it has proven effective within our design space and aligns well with real-world benchmarks. Our objective is to find appropriate values for block size $C$, latent dimension $P$, and LFSR length $K$, such that the reconstruction error is minimized when the optimal seed is selected. More specifically, let $\hat{s^*}_{C,P,K}$ and $\hat{\mathbf{t}^*}_{C,P,K}$ denote the solutions obtained from Eq.~\ref{eq:min_error}. For an $M$-bit compression, we solve the following optimization problem:

\begin{equation} \label{eq}
    \begin{aligned}
        \mathbb{E}[\epsilon_{\min}] := & \underset{C, P, K}{\text{min}} \quad \mathbb{E}\big[\lVert \mathbf{w} - \mathbf{U}(\hat{s^*}_{C,P,K}) \hat{\mathbf{t}^*}_{C, P, K} \rVert_2^2\big], \\
        & \text{subject to:} \quad M C = K + 4 + 4P \quad \text{and} \quad C,K,P \in \mathcal{Z}^+,
    \end{aligned}
\end{equation}

where $\mathcal{Z}^+$ represents the set of all positive integers. Since the optimization problem in~\eqref{eq} is not analytically tractable, we numerically solved it by conducting a grid search over $C$, $P$, and $K$ constrained to positive integers and the given bit budget. Understanding the trade-offs among $C$, $P$, and $K$ is important for optimizing the approximation within a given bit budget. Each of these parameters influences the overall performance and contributes to minimizing the reconstruction error.

One critical trade-off is between the LFSR seed length $K$ and the latent dimension $P$. Increasing $K$ reduces the expected minimum error $\mathbb{E}[\epsilon_{\min}]$ by providing more opportunities to find a better projection. However, this comes at a cost: as $K$ increases, the number of required bits grows, reducing the available bit budget for $P$. The objective is to find an optimal $K$ that effectively lowers $\mathbb{E}[\epsilon_{\min}]$ without overly constraining $P$, as that could lead to a significant increase in the overall error $\mathbb{E}[\epsilon_{\min}]$. Similarly, increasing $P$ helps capture more of the energy of the vector $\mathbf{w}$, thus reducing $\mathbb{E}[\epsilon_{\min}]$. However, this also requires more bits, which may limit the value of $K$. The key is to strike a balance where enough energy is captured without overly sacrificing the exploration of better projections through $K$. Finally, increasing $C$ expands the total bit budget, allowing for larger values of both $P$ and $K$. However, this also increases the potential for higher error, as expanding the space may dilute the precision of projections.

Overall, $C$, $P$, and $K$ are interdependent, and optimizing these parameters requires a careful numerical exploration of different combinations. Based on this analysis, we selected the configurations shown in Table~\ref{tab:configs} for $M\!=\!3$ and $M\!=\!4$, which are used in the experiments reported next.

\begin{table}[h] 
\centering
\caption{Selected configurations of $C$, $P$, and $K$ for $M\!=\!3$ and $M\!=\!4$.}
\begin{tabular}{|c|c|c|c|}
\midrule
Bits per element $M$  & Block size $C$ & Latent dimension $P$ & LFSR seed length $K$ \\
\midrule
3 & 12 & 4 & 16 \\
\midrule
4 & 8 & 3 & 16 \\
\midrule
\end{tabular}

\label{tab:configs}
\end{table}

\section{Experiments}  \label{S:Experiments}

We apply Algorithm~\ref{Alg} across all weight blocks of pretrained LLMs to find the seeds and coefficients that minimize reconstruction error (Eq.~\ref{eq:min_error}). Using the configurations from Table~\ref{tab:configs}, we evaluate our compression methods in terms of accuracy and performance. Our experiments focus on Llama 2 and Llama 3 models \citep{touvron2023llama}, and unlike other methods, SeedLM does not require fine-tuning or calibration data while still achieving competitive results. We assess model quality using perplexity and accuracy, followed by performance analysis through FPGA-based matrix multiplication with low-level LFSR generation. This highlights the cost and performance benefits of SeedLM, especially in hardware-constrained environments.

\subsection{Accuracy Results} \label{ss:exp_acc}

To evaluate the quality of SeedLM, we measure perplexity on the WikiText-2 dataset \citep{merity2016pointer} and assess accuracy across various zero-shot tasks using the LM Evaluation Harness \citep{gao2021framework}\footnote{For all compression methods, we use LM Evaluation Harness v0.4.3 and the following task versions: arc-challenge=1.0, arc-easy=1.0, hellaswag=1.0, winogrande=1.0, boolq=2.0.}. We compare our method against established compression techniques such as AWQ \citep{lin2024awq}, OmniQuant \citep{shao2023omniquant}, and QuIP\# \citep{tseng2024quip}, using the official GitHub repositories for each baseline as of September~2024. A key strength of SeedLM is that it can operate entirely data-free, in contrast to other methods that require calibration data to achieve comparable results. For the baseline methods, we use the default calibration sets from their official repositories. Our experiments involve Llama 2 models (7B, 13B, 70B) and Llama 3 models (8B, 70B), tested with 3-bit and 4-bit representations. In the case of AWQ and OmniQuant, we use 4-bit integers with channel-wise scaling to avoid significantly increasing the bits per element beyond the allocated 3 or 4 bits (since a group size of 128 in these methods adds roughly 0.25 extra bits per parameter). For QuIP\# and OmniQuant, we ensure a fair comparison with SeedLM and AWQ by not performing fine-tuning on the quantized models. However, unlike SeedLM, all of them still require per-layer calibration, which relies on calibration data and activations, whereas SeedLM achieves its results without any data dependency.

\begin{minipage}{0.31\textwidth}
    To evaluate general language model performance, we measure perplexity on the WikiText-2 language modeling dataset, using 166 windows, each with a length of 2048 tokens, from the test data split. The results, as shown in Table~\ref{tab:wiki-manual}, illustrate a clear trade-off between compression level and model quality. In some cases, larger models subjected to aggressive compression even underperform smaller models with milder compression. SeedLM consistently outperforms state-of-the-art compression techniques, particularly at higher compression levels. Notably, SeedLM achieves these results without the need for any calibration data.

\end{minipage}
\hfill
\begin{minipage}{0.66\textwidth}
    \centering
    \captionof{table}{WikiText-2 perplexity results for 3- and 4-bit representation of Llama 2 and Llama 3 models with a sequence length of 2048. The notation x-yB refers to the Llama x model with yB parameters (e.g., 2-7B means Llama 2 with 7 billion parameters). Perplexity values above 100 are shown as inf. The best perplexity values are highlighted, and results that ran out of memory on our setup (8 A100 40GB GPUs) are marked as OOM.}
    \begin{tabular}{llccccc} 
\toprule
\textbf{Method} & \textbf{Bits} & \textbf{2-7B} & \textbf{2-13B} & \textbf{2-70B} & \textbf{3-8B} & \textbf{3-70B} \\
\midrule
Baseline & 16 & 5.5 & 4.9 & 3.3 & 6.1 & 2.9 \\
\midrule
SeedLM & 4 & \cellcolor{lightgray} 5.7 & \cellcolor{lightgray}5.1 & \cellcolor{lightgray}3.5 & \cellcolor{lightgray}7.0 & \cellcolor{lightgray}3.8 \\
OmniQuant & 4 & 6.1 & 5.2 & 3.7 & inf & inf \\
AWQ & 4 & 5.8 & \cellcolor{lightgray}5.1 & \cellcolor{lightgray}3.5 & 7.1 & 4.7 \\
QuIP\# & 4 & 6.5 & 5.3 & OOM & 7.6 & OOM \\
\midrule
SeedLM & 3 &\cellcolor{lightgray} 6.6 & \cellcolor{lightgray}5.8 & \cellcolor{lightgray}4.0 & \cellcolor{lightgray}10.1 & \cellcolor{lightgray}5.7 \\
OmniQuant & 3 & inf & 10.7 & 7.5 & inf & inf \\
AWQ & 3 & 15.6 & 6.5 & 4.4 & 11.8 & 11.6 \\
QuIP\# & 3 & 10.8 & 5.7 & OOM & \cellcolor{lightgray}10.1 & OOM \\
\bottomrule
\label{tab:wiki-manual}
\end{tabular}
\end{minipage}

\begin{table}[t!]
\centering
\caption{Performance comparison across different models and zero-shot tasks for $4$-bit and $3$-bit configurations. Results that ran out of memory in our setup ($8$ A100
40GB GPUs) are marked with OOM.}
\label{tab:combined_performance_comparison}
\small
\begin{tabular}{@{}llccccccc@{}}
\toprule
\textbf{Model} & \textbf{Method} & \textbf{Bits} & \textbf{ARC-Easy} & \textbf{ARC-Challenge} & \textbf{HellaSwag} & \textbf{WinoGrande} & \textbf{BoolQ} & \textbf{Mean}\\ 
\midrule
\multirow{7}{*}{\makecell{Llama 2\\7B}} & Baseline & 16 & 74.58 & 46.33 & 75.98 & 69.06 & 77.74 & 68.74 \\
\cdashline{2-9}
& SeedLM & 4 & 73.23 & 44.54 & 74.45 & 68.43  & 77.19 & 67.57 \\
& AWQ & 4 & 70.58 & 43.94 & 74.96 & 68.75 & 78.29 & 67.30 \\
& QuIP\# & 4 & 68.35 & 39.85 & 72.40 & 65.59 & 75.14 & 64.27 \\
& OmniQuant & 4 & 70.71 & 43.52 & 74.20 & 68.27 & 73.64 & 66.07 \\
\cdashline{2-9}
& SeedLM & 3 & 69.87 & 41.21 & 70.72 & 66.30 & 74.28 & 64.48 \\
& AWQ & 3 & 53.37 & 33.62 & 56.66 & 61.09 & 57.58 & 52.46 \\
& QuIP\# & 3 & 59.51 & 34.22 & 59.23 & 61.09 & 65.20 & 55.85 \\
& OmniQuant & 3 & 35.69 & 25.77 & 35.48 & 52.88 & 42.48 & 38.46 \\
\midrule
\multirow{7}{*}{\makecell{Llama 2\\ 13B}} & Baseline & 16 & 77.44 & 48.98 & 79.38 & 72.22 & 80.55 & 71.71 \\
\cdashline{2-9}
& SeedLM  & 4 & 76.98 & 49.83 & 78.54 & 72.77 & 79.20 & 71.46 \\
& AWQ & 4 & 77.44 & 49.32 & 78.57 & 71.90 & 78.47 & 71.14 \\
& QuIP\# & 4 & 74.24 & 45.48 & 77.17 & 71.27 & 79.51 & 69.53 \\
& OmniQuant & 4 & 76.18 & 47.95 & 78.10 & 72.14 & 81.77 & 71.23 \\
\cdashline{2-9} 
& SeedLM  & 3 & 72.85 & 45.39 & 74.50 & 71.35 & 78.81 & 68.58 \\
& AWQ & 3 & 70.58 & 45.14 & 72.72 & 64.96 & 72.45 & 65.17 \\
& QuIP\# & 3 & 73.48 & 45.14 & 74.92 & 69.06 & 79.60 & 68.44 \\
& OmniQuant & 3 & 55.85 & 34.47 & 59.54 & 53.04 & 63.39 & 53.26 \\
\midrule
\multirow{7}{*}{\makecell{Llama 2\\70B}} & Baseline & 16 & 80.98 & 57.25 & 83.81 & 77.98 & 83.70 & 76.74 \\
\cdashline{2-9}
& SeedLM  & 4 & 81.14 & 56.40 & 82.97 & 76.72 & 82.26 & 75.90 \\
& AWQ & 4 & 80.98 & 56.66 & 83.24 & 77.19 & 83.27 & 76.27 \\
& QuIP\# & 4 & OOM & OOM & OOM & OOM & OOM & OOM \\
& OmniQuant & 4 & 79.59 & 55.97 & 82.67 & 76.80 & 83.43 & 75.69 \\
\cdashline{2-9} 
& SeedLM  & 3 & 79.00 & 53.84 & 80.51 & 76.80 & 79.02 & 73.83 \\
& AWQ & 3 & 80.26 & 55.80 & 80.50 & 73.01 & 80.00 & 73.91 \\
& QuIP\# & 3 & OOM & OOM & OOM & OOM & OOM & OOM \\
& OmniQuant & 3 & 63.59 & 39.51 & 68.24 & 62.04 & 65.23 & 59.72 \\
\midrule
\multirow{7}{*}{\makecell{Llama 3\\8B}} & Baseline & 16 & 76.81 & 52.73 & 76.97 & 72.93 & 81.87 & 72.26 \\
\cdashline{2-9}
& SeedLM  & 4 & 76.52 & 49.74 & 76.61 & 72.93 & 80.76 & 71.31 \\
& AWQ & 4 & 74.49 & 51.54 & 78.03 & 73.09 & 80.40 & 71.51 \\
& QuIP\# & 4 & 72.39 & 46.93 & 75.93 & 71.82 & 79.24 & 69.26 \\
& OmniQuant & 4 & 73.95 & 47.78 & 73.42 & 69.69 & 71.99 & 67.37 \\

\cdashline{2-9} 
& SeedLM  & 3 & 67.21 & 41.55 & 68.34 & 69.22 & 67.61 & 62.79 \\
& AWQ & 3 & 64.90 & 40.19 & 68.40 & 65.04 & 74.62 & 62.63 \\
& QuIP\# & 3 & 65.07 & 40.36 & 67.79 & 68.82 & 72.14 & 62.84 \\
& OmniQuant & 3 & 30.26 & 22.53 & 28.96 & 49.33 & 48.47 & 35.91 \\
\midrule
\multirow{7}{*}{\makecell{Llama 3\\70B}} & Baseline & 16 & 85.23 & 64.33 & 84.07 & 77.66 & 86.27 & 79.51 \\
\cdashline{2-9}
& SeedLM  & 4 & 83.80 & 59.30 & 83.84 & 77.74 & 85.60 & 78.06 \\
& AWQ & 4 & 80.98 & 57.94 & 82.84 & 60.54 & 79.39 & 72.34 \\
& QuIP\# & 4 & OOM & OOM & OOM & OOM & OOM & OOM \\
& OmniQuant & 4 & 25.13 & 26.54 & 26.36 & 51.38 & 37.83 & 33.45 \\
\cdashline{2-9} 
& SeedLM  & 3 & 78.45 & 52.22 & 80.77 & 77.35 & 84.59 & 74.68 \\
& AWQ & 3 & 65.87 & 45.14 & 70.76 & 55.88 & 69.08 & 61.35 \\
& QuIP\# & 3 & OOM & OOM & OOM & OOM & OOM & OOM \\
& OmniQuant & 3 & 25.21 & 25.94 & 26.15 & 49.64 & 37.83 & 32.95 \\
\bottomrule
\end{tabular}
\end{table}

Next, we show zero-shot accuracy across various tasks. As seen in Table~\ref{tab:combined_performance_comparison}, SeedLM performs on par with or better than state-of-the-art methods at the same bit rates. This underscores SeedLM's ability to maintain competitive accuracy without relying on calibration data. Note that Llama 3 is much more sensitive to compression than Llama 2, likely due to its more advanced architecture and significantly larger training dataset. Llama 3 was trained on $15$ trillion tokens, around seven times more than Llama 2's 2 trillion tokens, enabling it to capture more detailed language patterns and nuances.  This increased complexity and sensitivity to nuance make it less compressible without a noticeable drop in performance. However, as shown in Figure~\ref{fig:accuracy_compare}, SeedLM still significantly outperforms other state-of-the-art methods on Llama 3 in terms of Retained Accuracy (\%), i.e. ratio of the compressed model's accuracy to the full-precision FP16 model's accuracy. This demonstrates that SeedLM is not only effective for such a complex model but is also robust across various tasks, maintaining high accuracy.

\subsection{Performance Analysis} \label{ss:exp_perf}
In this section, we shift our focus from accuracy to performance analysis on hardware. Specifically, we explore how SeedLM can be efficiently implemented on an FPGA. FPGAs are ideal for this task because they allow for highly parallelized computations and can be reconfigured to handle specific workloads, making them well-suited for running compressed models with lower bit rates that are not well supported by conventional GPUs. With an FPGA, the LFSR generation can be supported at a low-level in hardware rather than relying on relatively expensive software implementations.

\begin{figure}[htbp] 
    \centering
        \includegraphics[width=0.8\linewidth]{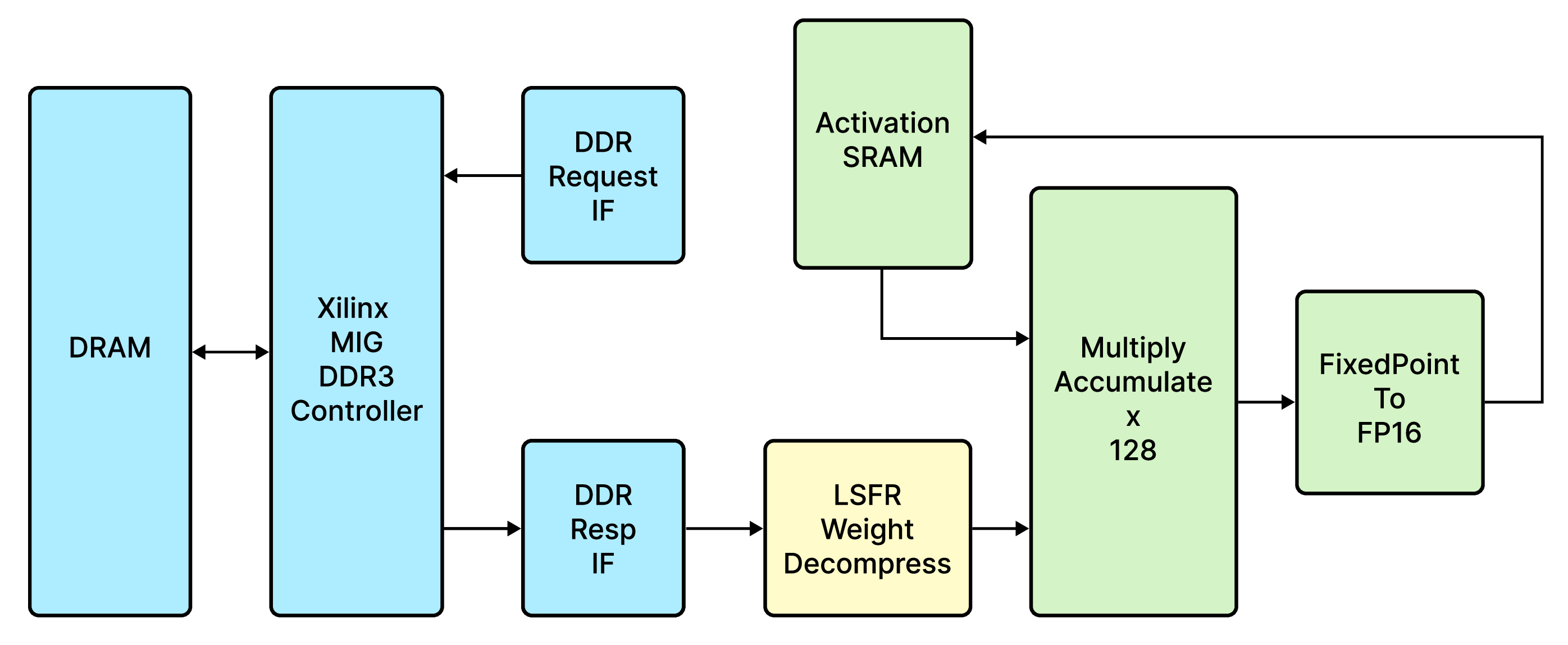}
    \caption{Block diagram of the RTL design.}
    \label{fig:FPGA}
\end{figure}

To evaluate SeedLM on an FPGA, we benchmark matrix-vector multiplication -- a core operation in most large language models -- both with and without SeedLM's 4-bit parametrization.  Figure~\ref{fig:FPGA} shows the RTL design block diagram, with the target device being an AMD Virtex7 FPGA~\cite{amdvirtex7datasheet}.

The implementation utilizes 128 DSP48 slice multipliers in parallel, calculating 128 elements of the activation vector simultaneously. The DDR response interface has a maximum bandwidth of 64 bytes per 200 MHz clock cycle, and the data path is designed to compute at the maximum DDR response throughput. When SeedLM compression is bypassed, the LFSR weight decompression is also skipped.

In the reference design, FP16 weights are read directly from DRAM, bypassing decompression. Due to the DDR interface’s 64-byte bus width, a maximum of 32 weight values can be read per cycle, limiting the utilization to only 32 of the 128 MACs. With SeedLM's 4-bit compression, however, 128 weight values can be read per cycle, resulting in a theoretical 4x performance improvement in memory access.

We use the reference implementation without compression (FP16 for all data) as a baseline to assess both performance and resource utilization on the FPGA. Table~\ref{table:FPGAresources} provides a detailed comparison of the FPGA resource usage, including LUTs, FFs, BlockRAM, and DSP counts.

\begin{itemize} \item \textbf{LUT} -- FPGA Fabric Lookup Table, used to create combinational logic. \item \textbf{FF} -- FPGA Fabric Register (Flip-Flop). \item \textbf{BRAM} -- FPGA Block RAM (SRAM), each BlockRAM is 36Kb. \item \textbf{DSPs} -- FPGA DSP48 resources, which include an 18-bit by 18-bit multiplier and accumulator. \end{itemize}

\begin{table}[h!]
\centering
\caption{FPGA resource utilization comparison}
\small
\begin{tabular}{|l|l|l|l|l|l|l|l|l|}
\toprule
\multirow{2}{*}{} & \multicolumn{2}{c}{\textbf{Total}} & \multicolumn{2}{c}{\textbf{MAC Block}} & \multicolumn{2}{c}{\textbf{LSFR Decompress}} & \multicolumn{2}{c|}{\textbf{PreMAC Fix2Float}} \\
\cmidrule{2-9}
 & \textbf{Reference} & \textbf{SeedLM} & \textbf{Reference} & \textbf{SeedLM} & \textbf{Reference} & \textbf{SeedLM} & \textbf{Reference} & \textbf{SeedLM} \\
\midrule
\textbf{LUTs} & 20800 & 120105 & 9902 & 42174 & 0 & 67034 & 0 & 12292 \\
\midrule
\textbf{FFs} & 10164 & 73666 & 3118 & 12594 & 0 & 45407 & 0 & 13448 \\
\midrule
\textbf{BRAMs} & 10.5 & 154.5 & 0 & 0 & 0 & 144 & 0 & 0 \\
\midrule
\textbf{DSPs} & 32 & 128 & 32 & 128 & 0 & 0 & 0 & 0 \\
\bottomrule
\end{tabular}
\label{table:FPGAresources}
\end{table}

In the reference design, only 32 MACs are used due to the input bandwidth limitation of 64 bytes per cycle. The SeedLM design, utilizing 128 MACs per cycle, results in approximately a 4x increase in MAC Block resources, aligning with the expected performance improvement.

Table~\ref{table:FPGAresources} reveals that SeedLM increases the total LUT count to 67K and register count to 45.4K, with the fixed-point to FP16 conversion accounting for 12.3K LUTs and 13.4K registers. The design includes 128 fully pipelined fixed-to-FP16 converters to sustain the DDR response data rate. We can note that performing all the compute in fixed-point math could eliminate the need for fixed-point to FP16 conversion, but this optimization is outside the scope of this paper. Table~\ref{table:FPGAcycles} shows the number of cycles required to complete matrix operations of various sizes, measured from the first DDR read to the final write to the activation SRAM. The SeedLM design achieves a 4x throughput compared to the reference design. As matrix size increases, the startup costs become less significant, leading the speedup to approach the theoretical 4x gain. In summary, SeedLM achieves near iso-accuracy at 4-bit compared to FP16, while offering close to a 4x speedup for memory-bound tasks such as generation in LLMs with billions of parameters and beyond.

\begin{table}[htbp]
\centering
\caption{Performance comparison for different matrix sizes}
\small
\begin{tabular}{|l|c|c|c|}
\toprule
 & \textbf{512 × 512} & \textbf{1024 × 1024
} & \textbf{2048 × 2048} \\
\midrule
\textbf{Reference} & 8593 & 34201 & 136559 \\
\midrule
\textbf{SeedLM} & 2341 & 8723 & 34331 \\
\midrule
\textbf{Speed Up} & 3.67 & 3.92 & 3.98 \\
\bottomrule
\end{tabular}
\label{table:FPGAcycles}
\end{table}

\section{Concluding Remarks}  \label{S:Conclusion}

In this paper, we presented SeedLM, a post-training compression method that uses pseudo-random generators to efficiently encode and compress model weights. SeedLM offers a data-free approach, avoiding the need for calibration data while retaining competitive accuracy, achieving up to around 98\% zero-shot accuracy at 3- and 4-bit quantization levels. We demonstrated the method's performance on both Llama 2 and Llama 3 models, showing that it performs comparably to existing state-of-the-art techniques. Furthermore, our FPGA implementation highlights SeedLM's potential for improved computational efficiency in hardware-constrained environments. While we believe that additional fine-tuning could further improve results we leave that to future works.

\subsubsection*{\textbf{Acknowledgments}}
\label{S:Ack}
We are grateful to Dmitry Belenko, Karen Khatamifard, Qingqing Cao, Seyed Mohsen Moosavi Dezfooli, and Arsalan Farooq for their invaluable feedback on this research.

\bibliography{iclr2025_conference}
\bibliographystyle{iclr2025_conference}

\newpage
\appendix
\section{Appendix}
\subsection{Coefficients Used in LFSRs} \label{appendix:lfsr_coeff}

The following table lists the indexed $j$ coefficients used for each $K$, where $\alpha_j$ equals one, with all other coefficients being zero. These specific coefficients correspond to the Linear Feedback Shift Registers (LFSRs) for each register length $K$, with the coefficients indexed starting from 0, representing the tap positions in the shift register. These coefficients are hard-wired into the hardware configuration of the LFSRs used in our experiments. These coefficients define the feedback polynomial for each LFSR, ensuring maximal-length cycles.

\begin{table}[ht]
\centering
\caption{Indexed $j$ coefficients used in LFSRs for each register length $K$, where $\alpha_j = 1$ and all other coefficients are zero.}
\begin{tabular}{|c|c|}
\toprule
\textbf{Length of LFSR ($k$)} & \textbf{Indices ($j$)} \\
\midrule
2  & (0, 1) \\
3  & (0, 1) \\
4  & (0, 1) \\
5  & (0, 2) \\
6  & (0, 1) \\
7  & (0, 1) \\
8  & (0, 2, 3, 4) \\
9  & (0, 4) \\
10 & (0, 3) \\
11 & (0, 2) \\
12 & (0, 1, 2, 8) \\
13 & (0, 1, 2, 5) \\
14 & (0, 1, 2, 12) \\
15 & (0, 1) \\
16 & (0, 1, 3, 12) \\
17 & (0, 3) \\
18 & (0, 7) \\
19 & (0, 1, 2, 5) \\
20 & (0, 3) \\
21 & (0, 2) \\
22 & (0, 1) \\
23 & (0, 5) \\
24 & (0, 1, 2, 7) \\
\bottomrule
\end{tabular}
\end{table}

\end{document}